\pdfoutput=1

\documentclass[11pt]{article}

\usepackage{acl}
\usepackage{times}
\usepackage{latexsym}

\usepackage[T1]{fontenc}

\usepackage[utf8]{inputenc}

\usepackage{microtype}

\usepackage{pifont}
\usepackage{bbding}
\usepackage{amsfonts}
\usepackage{amsmath}
\usepackage{booktabs}
\usepackage{multirow}
\usepackage{graphicx}
\usepackage{enumitem}
\usepackage{textcomp}
\usepackage{booktabs}
\usepackage{subfigure}
\usepackage{array}
\usepackage{graphicx}
\usepackage{caption}
\usepackage{hyperref}

\title{Dial-MAE: ConTextual Masked Auto-Encoder for Retrieval-based Dialogue Systems}

\author{
  Zhenpeng Su,
  Xing Wu,
  Wei Zhou,
  Guangyuan Ma,
  Songlin Hu
}

\author{Zhenpeng Su\textsuperscript{\rm 1,2}\footnotemark[1], Xing Wu\textsuperscript{\rm 1,2}\footnotemark[1], Wei Zhou\textsuperscript{\rm 1,2}\footnotemark[2], Guangyuan Ma\textsuperscript{\rm 1,2}, Songlin Hu\textsuperscript{\rm 1,2}\footnotemark[2] \\
  \textsuperscript{\rm 1}Institute of Information Engineering, Chinese Academy of Sciences, Beijing, China\\
  \textsuperscript{\rm 2}School of Cyber Security, University of Chinese Academy of Sciences, Beijing, China\\
  \texttt{\{suzhenpeng,wuxing,zhouwei,maguangyuan,husonglin\}@iie.ac.cn} \\}

\begin{document}
\maketitle
\begin{abstract}
Dialogue response selection aims to select an appropriate response from several candidates based on a given user and system utterance history. Most existing works primarily focus on post-training and fine-tuning tailored for cross-encoders. However, there are no post-training methods tailored for dense encoders in dialogue response selection. We argue that when the current language model, based on dense dialogue systems (such as BERT), is employed as a dense encoder, it separately encodes dialogue context and response, leading to a struggle to achieve the alignment of both representations. Thus, we propose Dial-MAE (Dialogue Contextual Masking Auto-Encoder), a straightforward yet effective post-training technique tailored for dense encoders in dialogue response selection. Dial-MAE uses an asymmetric encoder-decoder architecture to compress the dialogue semantics into dense vectors, which achieves better alignment between the features of the dialogue context and response. Our experiments have demonstrated that Dial-MAE is highly effective, achieving state-of-the-art performance on two commonly evaluated benchmarks.

\end{abstract}

\section{Introduction}
The retrieval-based dialogue system is a popular research topic. Pre-trained language models (PLMs), especially deep bidirectional Transformer Language Models (LMs) like BERT encoder ~\cite{DBLP:conf/nips/VaswaniSPUJGKP17,DBLP:conf/naacl/DevlinCLT19}, have been widely used in dialogue response. Common uses of deep LM are cross-encoder and bi-encoder~\cite{DBLP:conf/emnlp/GaoC21}.
Recent works ~\cite{DBLP:conf/cikm/GuLLLSWZ20,DBLP:conf/aaai/WhangLOLHLL21,DBLP:conf/aaai/XuTJZ0021,DBLP:conf/naacl/HanHKKS21,DBLP:journals/corr/abs-2203-00793} on dialogue response retrieval systems are mostly based on cross-encoders, which feed both the dialogue context and response directly into LM and use attention over all tokens to output a relevance score. Although cross-encoders have relatively stronger performances, they need to compute the matches for every possible combination of context-response pairs, which is time-consuming~\cite{DBLP:journals/corr/abs-2110-06612}. In practice, cross-encoders are often used for re-ranking after dialogue retrieval. In contrast, another common use of deep LM is the dense encoder, i.e. bi-encoder, which encodes dialogue context and response into the context vector and response vector respectively. The correlations between context and responses are computed using cosine similarity or dot product functions in vector space ~\cite{DBLP:journals/corr/abs-2110-06612,DBLP:journals/corr/abs-2203-05765}. The bi-encoders have a faster computational speed but usually perform worse than the cross-encoder. 

Bi-encoders generally underperform compared to cross-encoders due to two main reasons below~\cite{DBLP:conf/naacl/HanHKKS21,DBLP:conf/emnlp/GaoC21,DBLP:journals/corr/abs-2110-06612}. Firstly, bi-encoders encode dialogue context and responses separately, which lacks deep interaction like the cross-encoder~\cite{DBLP:conf/naacl/HanHKKS21}. We consider this as a potential information barrier that hinders the performance of bi-encoders, resulting in significant differences between the dense vector representations of the dialogue context and response vectors. Secondly, language models like BERT~\cite{DBLP:conf/naacl/DevlinCLT19} have not been trained to aggregate complex information into a single dense representation~\cite{DBLP:conf/emnlp/GaoC21}. Although using contrastive learning during the fine-tuning can alleviate the above two issues~\cite{DBLP:journals/corr/abs-2110-06612}, the discussion regarding their mitigation with post-training remains absent in dialogue response selection. We argue that post-training a PLM specifically tailored for the dense dialogue retrieval is essential for achieving optimal performance.

\begin{figure*}[h]
    \centering
    \includegraphics[width=0.9\textwidth]{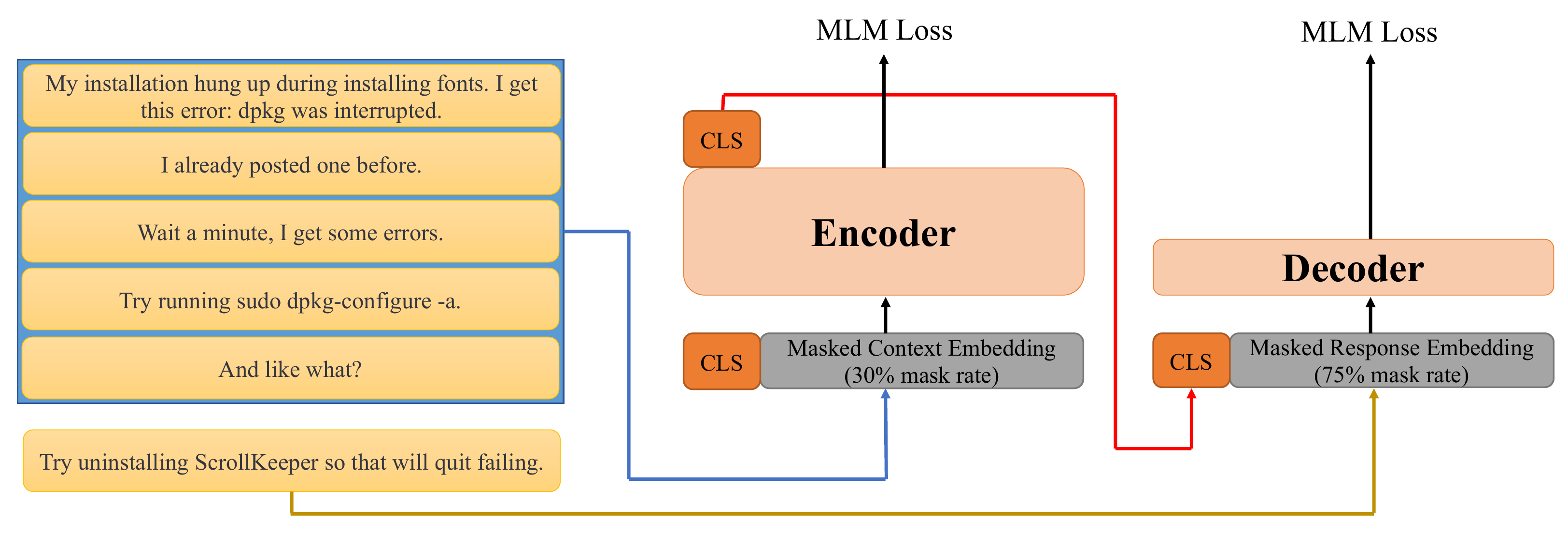}
    \caption{The model design for Dial-MAE. The input of the encoder is the dialogue context, and its next response and dialogue context embedding output by the encoder is used as the input to the decoder.}
    \label{fig:post-training}
\end{figure*}

In this paper, we focus on the above two issues and propose \textbf{Dial-MAE} (\textbf{Dial}ogue Contextual \textbf{M}asking \textbf{A}uto-\textbf{E}ncoder), a simple and effective post-training method tailored for the bi-encoder to compress dialogue semantic information and enhance the representation of dialogue-dense vectors. Our method provides a stronger foundation for model fine-tuning. 
Specifically, during the modeling process, we consider both the semantics of the dialogue context and the semantic relevance of the response. 

As shown in Figure \ref{fig:post-training}, we introduce an asymmetric encoder-decoder architecture. With the help of the dialogue context embedding $\mathbf{[CLS]}$ output by the encoder, the auxiliary task utilizes a weak decoder to reconstruct the masked response text. In other words, we employ the embedding of the dialogue context to directly generate responses. Therefore, even if the encoder side only receives the inputs of dialogue contexts, the output dialogue context embedding still needs to consider the correct response. This enables the dialogue context embedding $\mathbf{[CLS]}$ to incorporate contextual information. In addition, the encoder is required to directly predict the correct response when encoding the dialogue context, which breaks the information barrier between the context and the response. Therefore, the context and response features output by Dial-MAE are more similar, and our ablation experiments also prove this.

Furthermore, it is noteworthy that, similar to ~\cite{DBLP:conf/emnlp/XiaoLSC22,DBLP:conf/emnlp/GaoC21}, we apply asymmetric mask rates to the encoder and decoder. The decoder side has a higher mask rate than the encoder side. Such a design has the following advantage. Since the decoder has limited modeling capacity and high mask rate, the reconstruction on the decoder side is difficult to accomplish only by relying on masked response and rely more on the dialogue embedding output by the encoder, this forces the encoder to sufficiently aggregate the semantics of the dialogue context to aid the decoder in its MLM task~\cite{DBLP:conf/emnlp/XiaoLSC22,DBLP:conf/emnlp/GaoC21}.

Our contributions are as follows:
\begin{enumerate}
    \item We introduce Dial-MAE, a novel post-training method designed for bi-encoders, which utilizes dialogue context embeddings to generate responses, aiming to achieve feature alignment.
    \item We design a novel asymmetric encoder-decoder architecture to enhance the representational power of dialogue embedding.
    \item Experimental results show that in dialogue response retrieval, our method achieves state-of-the-art on two benchmarks with faster response speed.
\end{enumerate}

\section{Related Work}
In this section, we first discuss traditional retrieval dialogue systems based on neural networks, and then we discuss current dialogue systems based on pre-trained language models.
\subsection{Neural Dialogue Response Retrieval}

Dialogue response selection aims to select the most appropriate response from a range of candidates. Earlier studies  \cite{DBLP:journals/corr/KadlecSK15,DBLP:conf/sigdial/LowePSP15} focused on single-turn response selection. Later, more and more studies paid attention to multi-turn dialogue response selection.
\citeauthor{DBLP:conf/sigdial/LowePSP15} (\citeyear{DBLP:conf/sigdial/LowePSP15}) introduce a method that calculates the matching degree between dialogue and response based on Recurrent Neural Networks (RNNs). They also contributed a benchmark dataset named Ubuntu V1. 
In a similar vein, \citeauthor{DBLP:journals/corr/KadlecSK15} (\citeyear{DBLP:journals/corr/KadlecSK15}) advocate for the use of Convolutional Neural Networks (CNN) and Long Short-Term Memory (LSTM) as encoders to represent both the context and response. 
However, these methods do not explicitly treat each utterance as a unit, making it difficult to capture utterance-level discourse information.
\citeauthor{DBLP:conf/emnlp/ZhouDWZYTLY16} (\citeyear{DBLP:conf/emnlp/ZhouDWZYTLY16}) propose a multi-view model that encodes both word-level and utterance-level representations. Meanwhile, to fully reflect the relationship between dialogue and response, \citeauthor{DBLP:conf/acl/WuWXZL17} (\citeyear{DBLP:conf/acl/WuWXZL17}) suggest utilizing word embeddings and their sequential representations, encoded by Gated Recurrent Units (GRU), to construct a matching matrix between the dialogue context and response. 
With the popularity of attention mechanisms\cite{DBLP:conf/emnlp/LuongPM15,DBLP:conf/nips/VaswaniSPUJGKP17}. \citeauthor{zhou-etal-2018-multi} (\citeyear{zhou-etal-2018-multi}) propose a deep attention-matching network that applies the attention mechanism to the response selection dialogue system. 
Furthermore, \citeauthor{tao-etal-2019-one} (\citeyear{tao-etal-2019-one}) advocate for context and response matching by stacking multiple interaction blocks, providing a nuanced perspective. In a similar vein, \citeauthor{yuan-etal-2019-multi} (\citeyear{yuan-etal-2019-multi}) introduce a multi-hop selector network designed to identify relevant utterances in the context of response matching.
However, most traditional retrieval models are lightweight networks, and their performance is difficult to compare with PLMs.
\subsection{PLM-based Dialogue Response Retrieval }
Since PLMs show impressive performances in various downstream NLP tasks\cite{DBLP:conf/naacl/DevlinCLT19,DBLP:conf/nips/BrownMRSKDNSSAA20,DBLP:journals/corr/abs-2302-13971,su2023infoentropy,DBLP:conf/aaai/0002MLLWH23}. More and more studies apply PLMs to response selection. BERT-VFT \cite{DBLP:conf/interspeech/WhangLLYOL20} first applies the pre-trained language model BERT to dialogue response selection, and achieves state-of-the-art results. SA-BERT \cite{DBLP:conf/cikm/GuLLLSWZ20} adds speaker embedding to the model, in order to make the model aware of the speaker change information. Multi-Task Learning is also an effective way, UMS$_{BERT+}$ \cite{DBLP:conf/aaai/WhangLOLHLL21} proposes a set of strategies, which aids the response selection model towards maintaining dialogue coherence. Alternatively, \citeauthor{DBLP:conf/aaai/XuTJZ0021} (\citeyear{DBLP:conf/aaai/XuTJZ0021}) propose learning a context-response matching model with multiple auxiliary self-supervised tasks. 
However, these methods have the problem of not fully considering the relationship between each utterance in the context.
BERT-FP \cite{DBLP:conf/naacl/HanHKKS21} proposes to classify the relationship between a given utterance and a target utterance into more fine-grained labels, which makes the model learn the semantic relevance and coherence between the utterances. \citeauthor{DBLP:journals/corr/abs-2203-00793} (\citeyear{DBLP:journals/corr/abs-2203-00793}) propose two-level supervised contrastive learning so that the learned dialogue representations can be further separated in the embedding space.
In addition, DR-BERT\cite{DBLP:journals/corr/abs-2110-06612} explores the transfer of techniques from dense passage retrieval community to dialogue response selection. 
Although DR-BERT \cite{DBLP:journals/corr/abs-2110-06612} propose fine-tuning PLMs through contrastive learning to enhance the representation capability of dialogue-dense vectors, there has been no research on tailoring post-training tasks to enhance the representation ability of dialogue-dense vectors.

\section{Methodology}
This section first introduces masked language model pre-training as preliminary knowledge. Then we introduce detailed post-training, including the construction of data and the auxiliary task. Finally, we introduce the details of fine-tuning.
\subsection{Masked Language Model Pre-training}
MLM is an unsupervised method that masks parts of the input tokens and requires the Transformers-based LM to predict them based on the unmasked tokens. Formally, given an input sentence $X=[x_{1},x_{2},...,x_{n} ]$. We select a certain percentage of tokens from $X$ and replace them with a special token $\mathbf{[MASK]}$ to get corrupted $\widetilde{X}$. We denote these tokens replaced by $\mathbf{[MASK]}$ as $m(X)$. Then, LM is used to transform the corrupted input into the hidden states:
\begin{equation}
\begin{aligned}
    [\mathbf{h}_{cls}^{l}  ,\mathbf{h}_{}^{l} ]=LM([CLS],\tilde{X} )
\end{aligned}
\end{equation}
Here, $\mathbf{[CLS]}$ is a special token that is prepended at the beginning of the text. $\mathbf{h}_{cls}^{l}$ and $\mathbf{h}_{}^{l}$ respectively represent the hidden states of the final layer output after the $\mathbf{[CLS]}$ and $\widetilde{X}$ pass through the LM, i.e., $\mathbf{h}_{}^{l}$ = [$\mathbf{h}_{1}^{l}$,$\mathbf{h}_{2}^{l}$,...,$\mathbf{h}_{n}^{l}$]. For masked token, its corresponding hidden feature is used to predict the actual label. We formulate this process as:

\begin{equation}
\begin{aligned}
    \mathcal{L}_{mlm}=-\sum_{x_{i} \in m(X) }^{}  \log_{}{p(x_{i}|LM([CLS],\tilde{X} ))} 
\end{aligned}
\end{equation}

\subsection{Dial-MAE: Dialogue Contextual Masking Auto-Encoder}
Dial-MAE learns dialogue context information, which jointly models the semantics of the tokens inside a dialogue context and its response. 
We first describe how to build training data from all utterances of the dialogue session and then introduce the Dial-MAE post-training method. We randomly sample multiple consecutive utterances as context and the next utterance as its response. Multiple utterances of the context are connected using $\mathbf{[SEG]}$. For each dialogue scene, we sample multiple sets of such context and response pairs. The sampled context and response will serve as input to the encoder and decoder, respectively.

Then, we introduce the post-training design for Dial-MAE, as shown in Figure \ref{fig:post-training}, we use an asymmetric encoder-decoder: A deep encoder to generate dialogue context embedding, and a shallow transformer-based decoder (e.g. one or two layers) for response reconstruction. We apply a BERT encoder $Enc(.)$ with 12 layers, which receives masked dialogue context as inputs. The deep encoder has enough parameters to learn good dialogue representations, following the common practice, we select the final hidden state from the $\mathbf{[CLS]}$ token as the dialogue context embedding. The decoder is designed to assist the encoder in learning a better semantic representation of the dialogue. The input of the decoder $Dec(.)$ is the masked response as well as the dialogue context embedding, and it reconstructs the masked response tokens with the help of the context embedding. 

Through our design, the encoder $Enc(.)$ needs to predict the features of the correct response when encoding the dialogue context. This makes the dense encoder with behavior similar to that of a cross-encoder: simultaneously considering both the dialogue context and the response.
The advantage of doing this is to achieve feature alignment between the dialogue context and response during the post-training. Meanwhile, since the auxiliary MLM task breaks down the information barrier between separately encoding the dialogue context and response, the encoded output's $\mathbf{[CLS]}$ hidden state encompasses information from both.
Furthermore, it is worth noting that we employ an asymmetric masking operation(eg., 30\% for encoder, 75\% for decoder). On the decoder side, an aggressive mask rate and fewer model parameters will force its MLM task to rely more on the encoder's context embedding, which helps the encoder side aggregate complex information about the dialogue context into a dense vector.

Formally, we denote the dialogue context as $c$ and the response as $r$. We apply random mask operation to context to get $\tilde{c}$, denoting these tokens replaced by $\mathbf{[MASK]}$ in context as $m_{enc}(c)$. Similarly, we apply a random masking operation with a higher masking ratio for response to get $\tilde{r}$, denoting these tokens replaced by $\mathbf{[MASK]}$ in response as $m_{dec}(r)$. The encoding process can be expressed as:
\begin{equation}
\begin{aligned}
    [\mathbf{h}_{cls}^{c}  ,\mathbf{h}_{}^{c} ]=Enc([CLS],\tilde{c} )
\end{aligned}
\end{equation}
\begin{equation}
\begin{aligned}
    [\mathbf{h}_{cls}^{r}  ,\mathbf{h}_{}^{r} ]=Dec(\mathbf{h}_{cls}^{c},\tilde{r} )
\end{aligned}
\end{equation}
On the encoder side, the original context is learned to be reconstructed by optimizing the cross-entropy loss:
\begin{equation}
\begin{aligned}
    \mathcal{L}_{enc}=-\sum_{c_{i} \in m_{enc}(c) }^{}  \log_{}{p(c_{i}|Enc([CLS],\tilde{c} ))} 
\end{aligned}
\end{equation}
Differently, on the decoder side, the decoder reconstructs the original response with the help of the context embedding $h_{cls}^{c}$. We formulate this process as: 
\begin{equation}
\begin{aligned}
    \mathcal{L}_{dec}=-\sum_{r_{i} \in m_{dec}(r) }^{}  \log_{}{p(r_{i}|Dec(h_{cls}^{c},\tilde{r} ))} 
\end{aligned}
\end{equation}
Then, we add the encoder and decoder losses to obtain a summed loss:
\begin{equation}
\begin{aligned}
    \mathcal{L}= \mathcal{L}_{enc}+\mathcal{L}_{dec}
\end{aligned}
\end{equation}

\subsection{Fine-tuning for dialogue response selection}
\begin{figure}[h]
    \centering
    \includegraphics[width=0.5\textwidth]{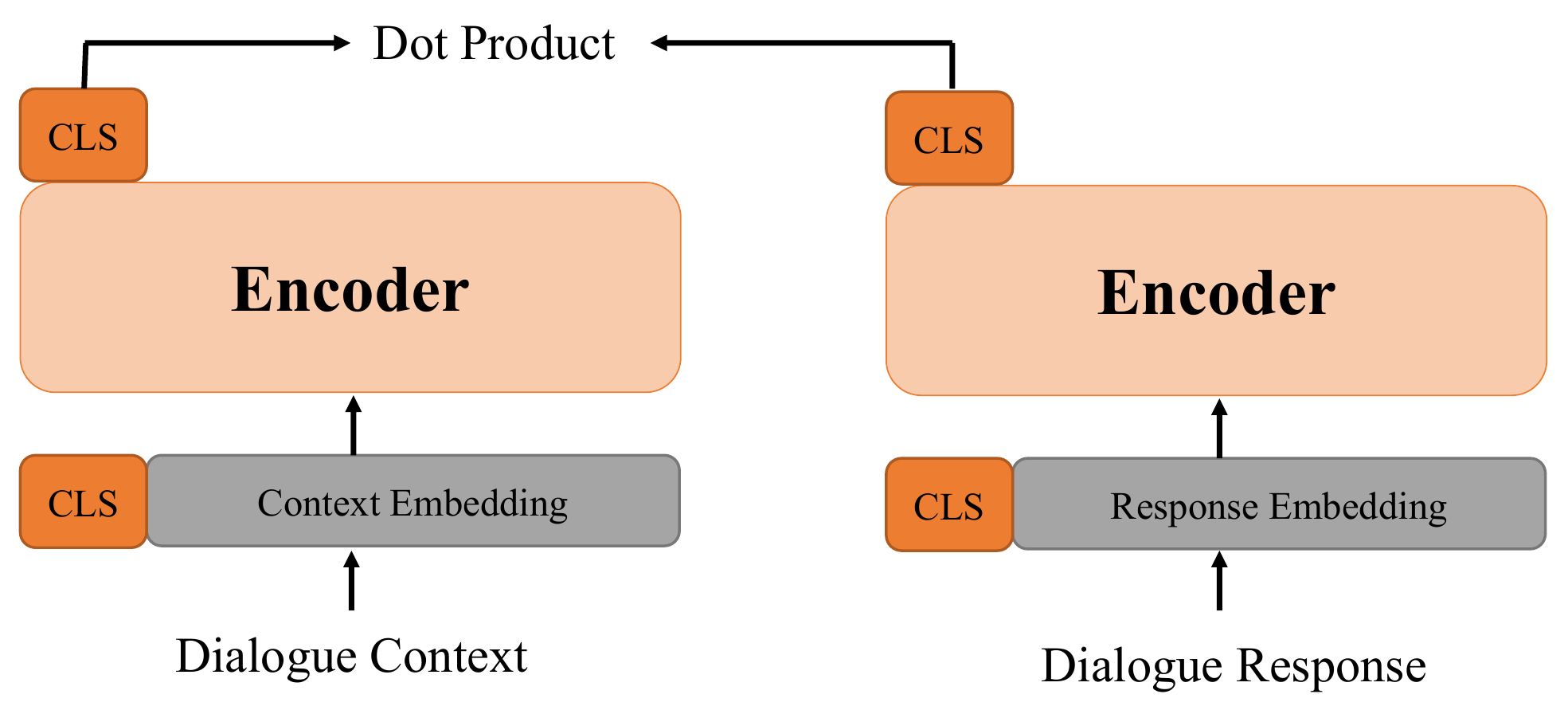}
    \caption{We discard the decoder, initialize the context encoder and response encoder using the encoder part of Dial-MAE, and fine-tune using contrastive learning. At inference time, We use a dot product to measure similarity.}
    \label{fig:finetune}
\end{figure}

At the end of Dial-MAE post-training, fine-tuning is conducted on the downstream dialogue response selection to verify the effectiveness of post-training. 
As shown in Figure \ref{fig:finetune}, in the fine-tuning stage, we only keep the encoder and discard the decoder. The encoder weights are used to initialize a dialogue context encoder $f_{c}$ and a response encoder $f_{r}$, respectively. 

The dialogue consists of a context $c$ that includes multiple utterances and a response $r$ with one utterance. After the dialogue context and response pass through the encoder, the context vector and response vector are respectively output. We train a dialogue response selection model using a contrastive learning loss function. 
\begin{equation}
\begin{aligned}
    \mathcal{L}_{ft}=\frac{exp(d(c,r^{+} ))}{exp(d(c,r^{+} ))+\sum_{j}^{}   exp(d(c,r_{j}^{-}  )) } 
\end{aligned}
\end{equation}
$r^{+}$ is the correct response corresponding to the dialogue context $c$. $r^{-}$ represents negative samples within a mini-batch. At inference time, we use the dot product $d(c,r)$ to measure the similarity between the context vector and the response vector:
\begin{equation}
\begin{aligned}
    d(c,r)=f_{c}(c)\cdot  f_{r}(r)
\end{aligned}
\end{equation}

\section{Experiment}
In this section, we first introduce our experimental details, including datasets, evaluation metrics, post-training, and fine-tuning. Then we introduce the experimental results.
\subsection{Datasets}
We tested our model on widely used benchmarks that include Ubuntu Corpus and E-commerce Corpus. The statistics for the two datasets are presented in Table \ref{data}.

\begin{enumerate}
    \item \textbf{Ubuntu Corpus.} Ubuntu IRC Corpus V1 \cite{DBLP:conf/sigdial/LowePSP15} is a publicly available domain-specific dialogue dataset. Each set of conversations has two participants discussing how to troubleshoot Ubuntu systems.
    \item \textbf{E-commerce Corpus.} E-commerce Corpus \cite{zhang-etal-2018-modeling} comprises genuine conversations in Chinese between customers and customer service personnel, collected from Taobao, a Chinese e-commerce platform.
\end{enumerate}

\begin{table}[!ht]
\centering
\scalebox{0.65}{
\begin{tabular}{l|ccc|ccc}
\toprule  
\textbf{Dataset}& \multicolumn{3}{c|}{Ubuntu}& \multicolumn{3}{c}{E-commerce}  \\
& train & val & test & train & val & test  \\
\midrule 
    \textbf{context-response pairs} & 1M  & 500k  & 500k  & 1M  & 10k  & 10k  \\
    \textbf{pos : neg} & 1:1  & 1:9  & 1:9  & 1:1  & 1:1  & 1:9 \\
    \textbf{avg turns} & 10.13  & 10.11  & 10.11  & 5.11  & 5.48  & 5.64  \\
\midrule
\end{tabular}
}
\caption{Statistics related to data for the Ubuntu and E-commerce Corpus.}
\label{data}
\end{table}

\subsection{Evaluation Metric}

We evaluated our model using $R_{10}@k$, following previous studies \cite{DBLP:conf/naacl/HanHKKS21,DBLP:journals/corr/abs-2203-00793}, we evaluate our model using $R_{10}@k$. The notation $R_{10}@k$ represents Recall, indicating that among ten possible responses, the correct answer is included within the top k options. 

\subsection{Implementation Details}
We first introduce the experimental setup for post-training, followed by the experimental setup for contrastive learning.

\paragraph{\textbf{Post-training.}} Dial-MAE's encoder is initialized with a pre-trained 12-layer BERT-base model, while the decoder is initialized from scratch. 
Specifically, following the previous works, for the E-commerce dataset, we employ bert-base-chinese\footnote{https://huggingface.co/bert-base-chinese}. For the Ubuntu dataset, we utilize the bert-base-uncased\footnote{https://huggingface.co/bert-base-uncased}.
We pre-train the model using the AdamW optimizer for a maximum of 15k steps, a global batch size of 1k, and a linear schedule with a warmup ratio of 0.1 on all two datasets. We set the input sequence lengths to 256 and 64 for the encoder and decoder, respectively. In fact, for the Chinese datasets E-commerce, we followed the parameter settings from Cot-MAE\cite{DBLP:conf/aaai/0002MLLWH23}: The masking ratio of the encoder is 30\%, the masking rate of the decoder is 45\%, the learning ratio is 1e-4, and the decoder has two layers.  Differently, for the English dataset Ubuntu, the masking ratio of the encoder is 30\%, the masking ratio of the decoder is 75\%, and the decoder is one layer. We also adjust the learning rate to 3e-4 to ensure the loss function converges. We set a widely used random seed as 42 for reproducibility. After post-training, we discard the decoder, only leaving the encoder for fine-tuning. 

\begin{table*}[!ht]
\centering
\scalebox{1}{
\begin{tabular}{l|ccc|ccc}
\toprule  
\textbf{Models}& \multicolumn{3}{c|}{Ubuntu} &  \multicolumn{3}{c}{E-commerce} \\
& $R_{10}@1$ & $R_{10}@2$ & $R_{10}@5$ & $R_{10}@1$ & $R_{10}@2$ & $R_{10}@5$ \\
\midrule 
    TF-IDF \cite{DBLP:conf/sigdial/LowePSP15} & 0.410  & 0.545  & 0.708  & 0.159  & 0.256  & 0.477 \\
    RNN \cite{DBLP:conf/sigdial/LowePSP15}  & 0.403  & 0.547  & 0.819 & 0.118  & 0.223  & 0.589 \\
    CNN \cite{DBLP:journals/corr/KadlecSK15}  & 0.549  & 0.684  & 0.896 & 0.328  & 0.515  & 0.792 \\
    LSTM \cite{DBLP:journals/corr/KadlecSK15}  & 0.638  & 0.784  & 0.949 & 0.365  & 0.536  & 0.828 \\
    SMN \cite{DBLP:conf/acl/WuWXZL17}  & 0.726  & 0.847  & 0.961  & 0.453  & 0.654 & 0.886 \\
    DUA \cite{zhang-etal-2018-modeling}  & 0.752  & 0.868  & 0.962 & 0.501  & 0.700  & 0.921 \\
    DAM \cite{zhou-etal-2018-multi}  & 0.767  & 0.874  & 0.969  & 0.526  & 0.727  & 0.933 \\
    IOI \cite{tao-etal-2019-one}  & 0.796  & 0.894  & 0.974 & 0.563  & 0.768  & 0.950 \\
    ESIM \cite{chen2019sequential}  & 0.796  & 0.894  & 0.975  & 0.570  & 0.767  & 0.948 \\
    MSN \cite{yuan-etal-2019-multi}  & 0.800  & 0.899  & 0.978  & 0.606  & 0.770  & 0.937 \\
    RoBERTa-SS-DA \cite{lu2020improving}  & 0.826  & 0.909  & 0.978 & 0.627  & 0.835  & 0.980 \\
    BERT-VFT \cite{DBLP:conf/interspeech/WhangLLYOL20}  & 0.855  & 0.928  & 0.985   & -  & -  & - \\
    SA-BERT \cite{DBLP:conf/cikm/GuLLLSWZ20}  & 0.855  & 0.928  & 0.983 & 0.704  & 0.879  & 0.985 \\
    UMS$_{BERT+}$ \textbf{\cite{DBLP:conf/aaai/WhangLOLHLL21}} & 0.875  & 0.942  & 0.988  & 0.764  & 0.905  & 0.986 \\
    BERT-SL \cite{DBLP:conf/aaai/XuTJZ0021}  & 0.884  & 0.946  & 0.990   & 0.776  & 0.919  & 0.991 \\
    DR-BERT \cite{DBLP:journals/corr/abs-2110-06612} $\clubsuit$ & 0.910 & 0.962 & 0.993  & - & - & - \\
    BERT-FP \cite{DBLP:conf/naacl/HanHKKS21} & \underline{0.911}  & 0.962  & \textbf{0.994}  & 0.870  & 0.956  & 0.993 \\
    BERT-TL \cite{DBLP:journals/corr/abs-2203-00793}  & 0.910  & \underline{0.962}  & 0.993  & \underline{0.927} &\underline{0.974} &\underline{0.997} \\
\midrule
    \textbf{BERT$_{+CL}$} & 0.887 & 0.948 & 0.989  & 0.849 & 0.937 & 0.991 \\
    \textbf{Dial-MAE} & \textbf{0.918}$^\ast $  & \textbf{0.964}$^\ast$   & \underline{0.993} & \textbf{0.930}$^\ast$  & \textbf{0.977}$^\ast$ & \textbf{0.997} \\
    diff. \%p & +3.1\% & +2.4\% & +0.4\% & +8.1\% &  +4\% & +0.6\% \\
\bottomrule
\end{tabular}
}
\caption{Main experiment results on E-commerce Corpus and Ubuntu Corpus. \textbf{BERT$_{+CL}$} means fine-tuning BERT using contrastive learning. The best score on a given dataset is marked in \textbf{bold}, and the second best is \underline{underlined}. $\clubsuit:$ According to the published code, for E-commerce, they adjusted the hyperparameters on the test set without cross-validation, we think the results are misleading, and this part has been removed. 
Two-tailed t-tests demonstrate statistically significant improvements of Dial-MAE over baselines ($\ast \le 0.01$).
}
\label{result}
\end{table*}

\paragraph{\textbf{Fine-tuning.}} We fine-tune using contrastive learning on each dataset. During training, we follow \cite{DBLP:journals/corr/abs-2110-06612} regarding every utterance in the dialogue sense as a response and its previous utterances as a context. Our model is optimized by AdamW optimizer, and the linear learning ratio scheduler is used. We tuned the hypermeters of individual tasks on their development sets. For Ubuntu, we fine-tune for 5 epochs, the learning rate is set to 5e-5, and the batch size is set to 64. For E-commerce, we fine-tune for 2 epochs, the learning rate is set to 1e-4, and the batch size is set to 128. We set a widely used random seed as 42 for reproducibility. 

\subsection{Results and Discussions}

We show the main results in Table \ref{result}, which shows that Dial-MAE achieves new state-of-the-art on the Ubuntu dataset and E-commerce dataset. We are able to achieve comparable performance to the state-of-the-art cross-encoders using a bi-encoder, and we have lower computational requirements compared to cross-encoders. Compared to BERT-FP, our model achieved an absolute improvement of 0.7\%p in $R_{10}@1$ on the Ubuntu Corpus and 6\%p in $R_{10}@1$ on the E-commerce. Compared to BERT-TL, our model achieves an absolute improvement of 0.8\%p in $R_{10}@1$ on the Ubuntu Corpus and a slight improvement of 0.3\%p in E-commerce. This suggests that our carefully tailored post-training method for the bi-encoder can achieve comparable performance to the complex-designed cross-encoder.

BERT$_{+CL}$ means fine-tuning BERT using contrastive learning. In comparison to BERT$_{+CL}$, Dial-MAE achieve an absolute improvement in $R_{10}@1$ by 3.1\%p, 8.1\%p on Ubuntu Corpus and E-commerce Corpus, respectively. This suggests that our custom post-training approach for dialogue retrieval models is effective. Aligning the features of the dialogue context and response during post-training enables improvements in contrastive fine-tuning. We believe the improvement comes from two aspects. On the one hand, the post-training method considers both the semantics of the tokens inside the context and its response. On the other hand, the asymmetric encoder-decoder structure with an asymmetric masking strategy facilitates post-training, which forces the encoder to learn better dialogue embeddings.

\begin{table*}[!ht]
\centering
\scalebox{0.9}{
\begin{tabular}{l|ccc|ccc}
\toprule  
\textbf{Models}& \multicolumn{3}{c|}{Ubuntu} &  \multicolumn{3}{c}{E-commerce} \\
& $R_{10}@1$ & $R_{10}@2$ & $R_{10}@5$ & $R_{10}@1$ & $R_{10}@2$ & $R_{10}@5$ \\
\midrule 
    \textbf{BERT$_{+CL}$} & 0.887 & 0.948 & 0.989 & 0.849 & 0.937 & 0.991 \\
    \textit{w/o Contrastive loss} & 0.205 & 0.341 & 0.647 & 0.141 & 0.242 & 0.466 \\
    \textbf{Dial-MAE} & \textbf{0.918}  & \textbf{0.964}  & \textbf{0.993} & \textbf{0.930} & \textbf{0.977} & \textbf{0.997} \\
    \textit{w/o Contrastive loss} & 0.783  & 0.867  & 0.950 & 0.483  & 0.639  & 0.853 \\
\midrule
\end{tabular}
}
\caption{Ablation results on the test sets of the two benchmarks.}
\label{ablation}
\end{table*}

\subsection{Ablation Study}
In this section, we analyze the experimental results to demonstrate
the effectiveness of the proposed Dial-MAE method. In the following experimental analysis, due to high computing budgets, most experiments use Ubuntu Corpus.

\paragraph{\textbf{The Impact of Auxilary Network.}} We remove the contrastive loss in BERT$_{+CL}$ and Dial-MAE, then evaluate their performance changes. As shown in Table \ref{ablation}, Dial-MAE achieved an absolute improvement in $R_{10}@1$ by 57.7\%p, and 34.2\%p on Ubuntu Corpus and E-commerce Corpus, respectively. 

This suggests that our proposed post-training method effectively achieves the alignment of contextual representations, making the dialogue context more similar to the features of the response.
We believe the gain comes from our auxiliary network helping the encoder aggregate dialogue contextual information. 
First, the encoder achieves feature alignment in the dialogue's contextual information by predicting the features of the correct response during the encoding of the context. Secondly, due to the small number of parameters of the decoder and the high mask rate on the decoder side, this will force the MLM task of the decoder to rely more on the dialogue context embedding output by the encoder. This enables the decoder to aggregate complex information about the dialogue context into a dense vector.

\begin{figure}[h]
    \centering
    \includegraphics[width=0.5\textwidth]{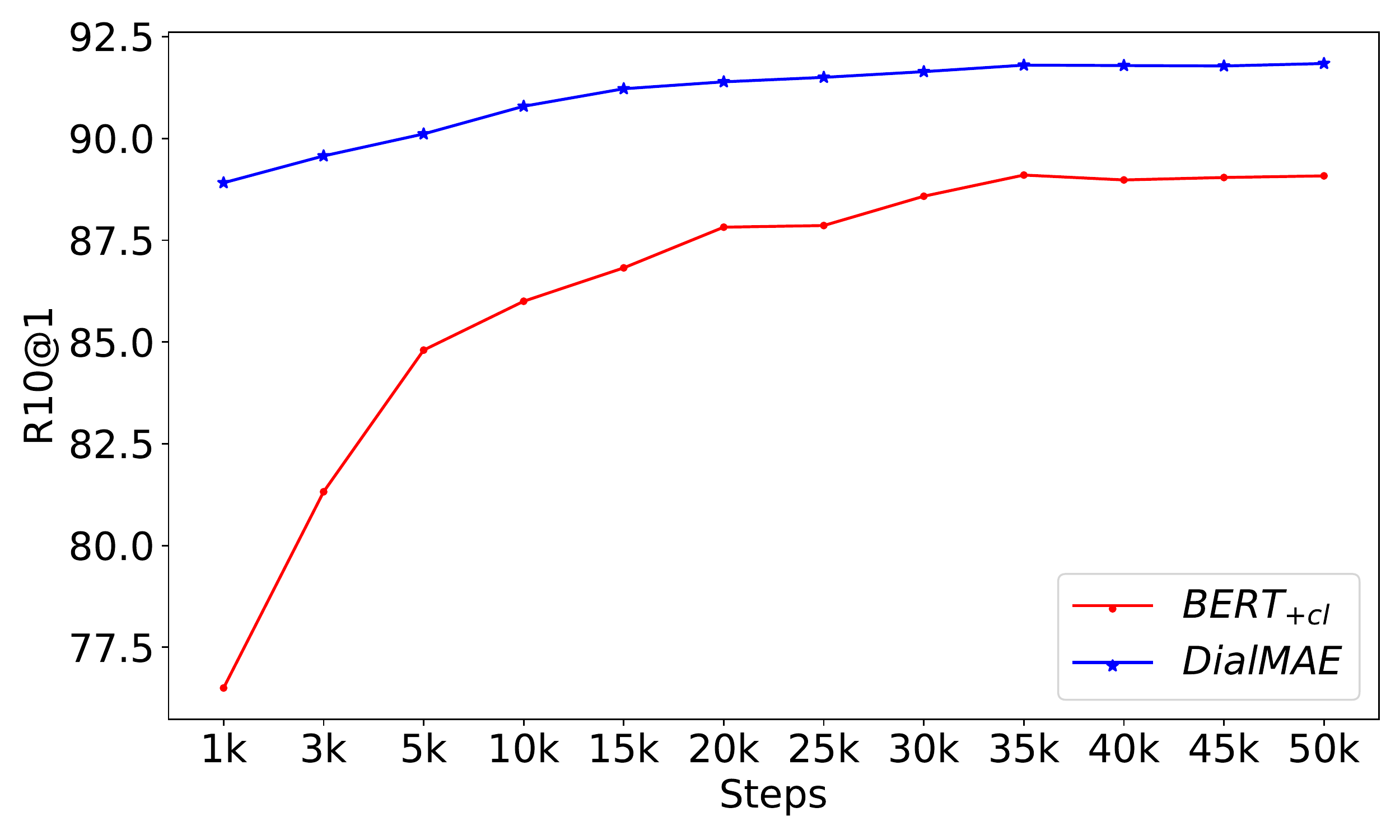}
    \caption{Fine-tuning schedules on the dev set of Ubuntu Corpus. A longer fine-tuning schedule gives a noticeable improvement. The performance of Dial-MAE is always better than BERT$_{+CL}$.}
    \label{fig:steps}
\end{figure}

We then use contrastive learning to fine-tune the post-training models, and the performance of the models can be further improved. We also give the fine-tuning schedule on Ubuntu Corpus as shown in Figure \ref{fig:steps}, with the accuracy steadily improving as the training time increases, and Dial-MAE consistently outperforms BERT$_{+CL}$. This result shows that both the contrastive loss and the auxiliary MLM loss are crucial in our method. Both contrastive learning and our post-training method are effective in achieving dialogue context and response feature alignment, and their effects can be additive. 

\begin{table}[!ht]
\centering
\scalebox{0.95}{
\begin{tabular}{lccl|lccc}
\toprule 
& Enc & Dec & & &  $R_{10}@1$ & $R_{10}@2$ & $R_{10}@5$\\
\midrule
& 0.15 & 0 & & & 91.0 & 96.0 & 99.2 \\
& 0.15 & 0.15 & & & 91.3 & 96.1 & 99.2 \\
& 0.15 & 0.45 & & & 91.5 & 96.2 & 99.3 \\
& 0.15 & 0.75 & & & 91.5 & 96.3 & 99.3 \\
& 0.30 & 0.45 & & & 91.7 & 96.5 & 99.3 \\
& 0.30 & 0.75 & & & \textbf{91.9} & \textbf{96.5} & 99.3 \\
& 0.30 & 0.90 & & & 91.6 & 96.4 & 99.3 \\
& 0.45 & 0.75 & & & 91.8 & 96.4  & \textbf{99.4} \\
\midrule
\end{tabular}
}
\caption{Impact of mask rate on the dev set of Ubuntu Corpus.  "Enc" denotes encoder, "Dec" denotes decoder. "Enc=0.15 Dec=0" means only using BERT's native MLM task without the decoder part.}\label{mask}
\end{table}

 \paragraph{\textbf{Impact of Mask Rate.}} 
 \citeauthor{DBLP:conf/aaai/0002MLLWH23} (\citeyear{DBLP:conf/aaai/0002MLLWH23}) find that using a larger mask rate in both the encoder and decoder can enhance the performance of the contextual masking Auto-Encoder. As shown in Table \ref{mask}, in our experiments, we find that an aggressive mask rate helps the learning of Dial-MAE. when the encoder mask rate equals 30\%, and the decoder mask rate equals 75\%, Dial-MAE achieves the best performance. 
When the encoder mask rate stays below 30\%, the performance of Dial-MAE improves as the decoder mask rate increases. When the encoder mask rate rises to 45\%, Dial-MAE's performance declines slightly. We believe this is due to the encoder doesn't provide enough dialogue context semantic information when its mask rate is too high. In addition, from the experimental results, no matter what set of mask rates, Dial-MAE obviously exceeds the result of post-training for MLM tasks alone, which proves the robustness of Dial-MAE.

\begin{figure}[h]
\centering
\includegraphics[width=0.5\textwidth]{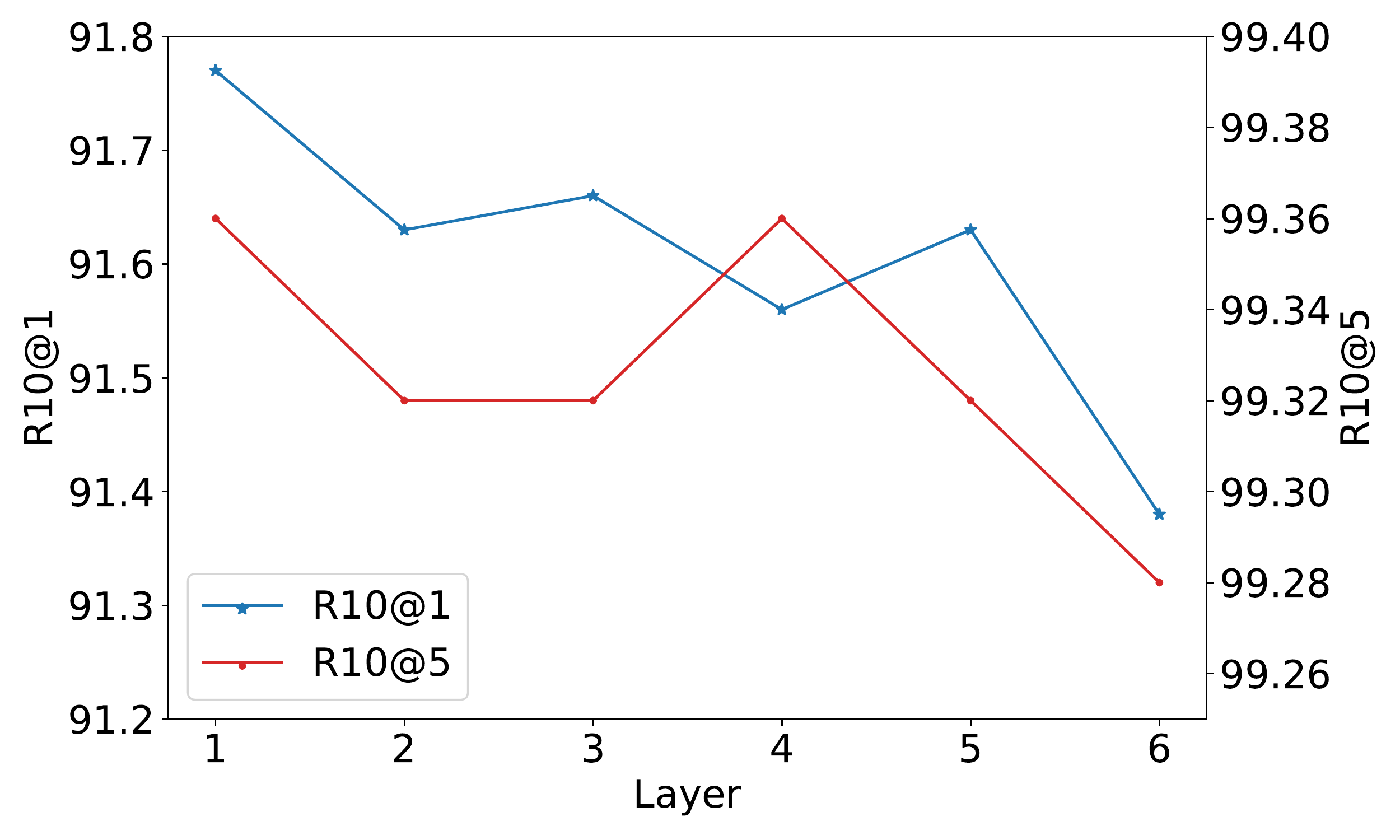}
\caption{Impact of layer number on Ubuntu Corpus.}
\label{fig:layer}
\end{figure}

\paragraph{\textbf{Impact of Decoder Layer Number.}} As shown in Figure \ref{fig:layer}, we further explore the impact of different decoder layer numbers on Dial-MAE performance. we find that using only one layer of the decoder yields the best results. Fewer decoder parameters can force the auxiliary MLM task to rely more on dialogue context embeddings output by the encoder. We believe that the more layers of the decoder, the stronger the decoding ability, and the decoder's dependence on context embedding will decrease, leading to insufficient constraints on encoder training. In general, no matter what set of layers, R@1 obviously exceeds the result of post-training for MLM tasks alone (Enc=0.15 Dec=0), as shown in Table \ref{mask}, which proves the robustness of Dial-MAE.

\paragraph{\textbf{Compared with Dense Models.}} To further illustrate the effectiveness of our custom approach for bi-encoders in dialogue response selection, we compared it with state-of-the-art dense models in the Information Retrieval(IR) community. On the Ubuntu dataset, we fine-tune the dense models proposed by the IR community using contrastive learning, and the experimental results are shown in the table \ref{compre_dense}. During pre-training, the corpus of CoT-MAE\cite{DBLP:conf/aaai/0002MLLWH23} and RetroMAE\cite{DBLP:conf/emnlp/XiaoLSC22} contains an additional 3.2M documents dataset MS-MARCO\cite{DBLP:conf/nips/NguyenRSGTMD16} in addition to BooksCorpus and Wikipedia. However, our experimental results show that although the results of the three dense models have improved compared with BERT$_{+CL}$, they are still not as good as our proposed Dial-MAE. This shows that our proposed method is better suited for encoding dense vectors of dialogue than other dense models.
\begin{table}[!ht]
\centering
\scalebox{0.7}{
\begin{tabular}{l|ccc}
\toprule  
\textbf{Models} & $R_{10}@1$ & $R_{10}@2$ & $R_{10}@5$ \\
\midrule 
    \textbf{BERT$_{+CL}$} & 89.2 & 95.1 & 99.2 \\
    \textbf{Condenser\cite{DBLP:conf/emnlp/GaoC21}} & 89.4 & 95.4 & 99.1 \\
    \textbf{RetroMAE\cite{DBLP:conf/emnlp/XiaoLSC22}} & 89.3 & 95.3 & 99.1 \\
    \textbf{Cot-MAE\cite{DBLP:conf/aaai/0002MLLWH23}} & 89.8 & 95.9 & 99.2 \\
\midrule
    \textbf{Dial-MAE} & \textbf{91.9} & \textbf{96.5} & \textbf{99.3} \\
\bottomrule
\end{tabular}
}
\caption{Comparison results of Dial-MAE and dense retrieval models on the Ubuntu dev set.}
\label{compre_dense}
\end{table}

\paragraph{\textbf{Qualitative Analysis.}}

To qualitatively analyze our post-training method, as shown in Table \ref{table_quality}, we provide the example. Dial-MAE can sort out the most appropriate response more accurately than BERT$_{+CL}$. The response sorted by BERT$_{+CL}$ have some token overlap with the dialogue context but are not semantically related. Compared with BERT$_{+CL}$, Dial-MAE can better understand dialogue semantics due to the joint modeling of context and response through post-training. This further demonstrates the effectiveness of our method.

\begin{table}[!htbp]
\centering
\small
\begin{tabular}{lcp{4cm}}
\toprule  
Relevant & Model  & Rank 1st response\\
\midrule
& & USER\_A:I already have these disks in my system just want to migrate my current homefolder to the new drive.

USER\_B:Mount your new disk to a temporarely mount point move or copy your home folder to the new disc after that delete you old homedir than mount the new disk to \_path\_. \\
\midrule
\multirow{1}{*}{\qquad \XSolidBrush} &\multirow{1}{*}{BERT$_{+CL}$} &  USER\_A: Do you use the ubuntu desktop or server i am using the desktop on my laptop i have only started with the server a little before \_number\_ came out maybe before april. \\
\midrule
\multirow{1}{*}{\qquad \Checkmark} & \multirow{1}{*}{Dial-MAE}  & USER\_A: I presume i use gparted to get the mountpoints correct or am i wrong \\
\bottomrule
\end{tabular}
\caption{Examples of rank 1st response recalled by different models on the the Ubuntu Corpus.}
\label{table_quality}
\end{table}

\section{Conclusion}
In this paper, we propose a post-training method tailored for dialogue response, considering the semantics of dialogue context and its corresponding responses. Precisely, we leverage a shallow decoder to force the encoder output dialogue embeddings to be more expressive. Experimental results show that our post-training method leads to considerable improvements, achieving state-of-the-art on two benchmark datasets. We also demonstrate the effectiveness of Dial-MAE through ablation experiments. Specifically, both contrastive learning and our post-training method are effective in achieving dialogue context and response feature alignment, and their effects can be additive. 

\section{Limitations}
Recently, generative conversational models based on large language models (LLMs) have demonstrated powerful performance. Despite the advantages of retrieval-based dialogue models in terms of computational cost and answer controllability, generative conversational systems based on LLMs surpass retrieval-based models in terms of answer diversity and flexibility. Furthermore, there has been much recent work exploring retrieval-augmented generation (RAG). In the future, we will further expand Dial-MAE to explore the effective integration with LLMs, using a dialogue response selection approach to attempt to address issues such as large model hallucinations and challenges related to knowledge updates. We hope that our work can also bring benefits to large language models.

\bibliography{anthology,custom}
\bibliographystyle{acl_natbib}

\end{document}